# Non-convex cost functionals in boosting algorithms and methods for panel selection

M. Visentin[*]

*mvisenti@itc.it*

**Abstract**

In this document we propose a new improvement for boosting techniques as proposed in ([2, 3]) by the use of non-convex cost functional. The idea is to introduce a correlation term to better deal with forecasting of additive time series. The problem is discussed in a theoretical way to prove the existence of minimizing sequence, and in a numerical way to propose a new *ArgMin* algorithm. The model has been used to perform the touristic presence forecast for the winter season 1999/2000 in Trentino (italian Alps).

## 1 Introduction

Here we study the possibility of the use of non-convex cost functionals to adapt boosting techniques in modeling and forecasting for additive time series. By an additive time series we mean the additive aggregation of a family $\mathcal{S} = \{s_n(t)\}_{n=1...N}$ of time series i.e. we need to model their pointwise sum:

$$f(t) = \sum_{n=1}^{N} s_n(t) \qquad (1)$$

Like in [2, 3] we want an additive model of (1) of the type:

$$\tilde{f}(t) = \sum_{m=0}^{M} \rho_m h(t, a_m) \qquad (2)$$

Under the assumption we will discuss later, to choose the $h(t, a_m)$ in $\mathcal{S}$ itself, we require in (2) that $M << N$. Further, we want to improve the predictive power of (2) introducing a *smoothing parameter* $\alpha$.

The problem of growing a model for (1) is solved in [2, 3] in many situations, but here we want to focus on the possibility to perform some forecasting analysis studying the behaviour of the metaparameters and the choice of the smoothing parameter.

---

[*]http://mpa.itc.it/mvisenti/home.html

Suppose now that your (continuous) time series $s_n \in \mathcal{S}$ are defined over an interval $I \subseteq \mathbb{R}$ (with functional analysis notation it means $\mathcal{S} \subset \mathcal{C}^0(I)$) and suppose that accordingly to some criteria you can partition $I$ into a training set $T$, a validation set $V$ to perform the prediction over new data (i.e. new time series) defined over a new interval $J$.

The algorithm actually computes only the parameters $a_m$ and $\rho_m$, for every fixed choice of the metaparameters, and one has to be establish them after the fitting procedure over the train set. The main metaparameters we have to discuss in the (2) are:

- the choice of the functions subset $\mathcal{F}(I \subseteq \mathbb{R})$ on which the algorithm operates;
- the best partition of the $I$;
- the number $M$ of function that we need to estimate the global presence function $f$, i.e. the dimension of the extracted panel;
- a value for additional smoothing coefficients for the $\tilde{f}$.

The interesting part of the problem is the choice of the metaparameters in a predictive way. At the and we give an application of this study to solve the problem of touristic presence. Suppose that you want to know on-line the behaviour of a touristic region where operates say $N$ hotels. So in these world the (1) is the global presence time series and its model (2) represents a panel extraction of $M$ hotels among the available $N$, with weights $\rho_i$ given to each selected hotel. Further if you see the (1) as a time series, at each time $t$ you are able to estimate also its integral that is the global presence in terms of units, up to the fixed $t$ day.

## 2 Minimization of a functional in $L^2$

Let consider a bounded interval $I$ of $\mathbb{R}$, and the Hilbert space $L^2(I)$. If:

$$<s,t>_{L^2} = \int_I s \cdot t \, dx$$

denotes the inner product of $L^2(I)$, let:

$$\rho(f,g) = \frac{<f-\bar{f}, g-\bar{g}>_{L^2}}{\sqrt{<f-\bar{f}, f-\bar{f}>_{L^2} \cdot <g-\bar{g}, g-\bar{g}>_{L^2}}} \quad (3)$$

where $\bar{f}$ is the mean of $f$ over $I$; the functional $\rho : L^2(I) \times L^2(I) \to [-1, 1]$ that reduces to Pearson's correlation coefficients in the discrete world.

Let $t(x)$ a convex function in $x$ such that $t(1) = 0$, and consider for a fixed $f \in L^2(I)$ the functional $\varphi_f(g) = t(\rho(f,g))$. In general $\varphi_f$ is no longer a convex functional, but with a good choice of $t(x)$, the $\varphi_f$ can be sequentially lower-semicontinuous, and coercive. For instance if $t(x) = \frac{1}{2+x} - \frac{1}{3}$ we have a convex function (over $[-1, 1]$) that takes its maximum in $x = -1$ and its minimum in $x = 1$. If we interpret these facts in terms of correlation, we can forget convexity and take $t(x) = \frac{1}{1+x^2} - \frac{1}{2}$. When $\rho$ can be interpreted as the correlation, this means that a good solution of the minimum problem is well correlated (or totally uncorrelated, depending on the choice of $t$) to the fixed $f$.

Now fix $f$ like in (1) and let:

$$\Psi(g) = \frac{1}{2} <f-g, f-g>_{L^2} + \varphi_f(g) \qquad (4)$$

the error functional $\Psi : L^2(I) \to \mathbb{R} \cup \{+\infty\}$ for a choice of $t(x)$ as above.

Note that the functional in general is not linear nor convex. The difference between the two functionals we proposed lies in the $\varphi$ part that carries just about the correlation of the solution of the problem. We think this correction is necessary because we are doing a linear additive model of an additive function. Now we discuss the possibilities to find a solution to the minimization problem for $\Psi$. First observe that for those choices of $t(x)$ such that $t(1) = 0$, $\Psi$ has a minimum over $L^2(I)$, that trivially is $f$. Before proceeding let's recall some definition about functional spaces:

**Definition 2.1** *A functional $J$ over a metric space $B$ is said to be coercive if and only if $J(x_n) \to +\infty$ whenever $\|x_n\| \to +\infty$.*

Observe that ours $\varphi_f$ are bounded and the $\Psi$ are coercive.

**Definition 2.2** *Let $B$ a Banach space and $J : B \to \mathbb{R} \cup \{+\infty\}$ a linear functional; $J$ is (sequentially) lower semicontinuous if and only if $J(x) \leq \liminf_n J(x_n)$ whenever $x_n \to x$ in $B$.*

Note that if $J$ is lower semicontinuous and $\{x_n\}$ is a minimizing sequence that is $J(x_n) \to \inf J$, and it is possible to extract a converging subsequence $x_{n_k} \to x$ then $\inf J \leq J(x) \leq \liminf J(x_{n_k}) = \inf J$. Moreover if $J$ is sequentially continuous, then is also lower semicontiuous.

**Proposition 2.3** *Let $\Psi$ a functional over $L^2(I)$ as defined in the (4) where $I$ is a bounded interval of $\mathbb{R}$, and $t(x)$ is continuous over $[-1, 1]$. Then $\Psi$ is sequentially continuous.*

*Proof:* Let $\{h_n\}_n \subseteq L^2(I)$ such that $h_n \to h$. Since $1 \in L^2(I)$, the means $\bar{h}_n \to \bar{h}$ (in $\mathbb{R}$) so we can assume that $\bar{h}_n = 0 \quad \forall n$. Observe that the sequence converges weakly, so the sequence $\{\|h_n\|\}_n$ is bounded in $\mathbb{R}$ so there exists a constant $C$ such that $|h_n| \leq C$ a.e. on I and by Lebesgue theorem it follows that $\varphi(h_n) \to \varphi(h)$, and then also the $\Psi$ are sequentially continuous. It remains to show the boundness property of the $\{h_n\}$. But the sequence converges weakly, so there exists a constant $K$ such that the sequence $\{\|h_n\|_{L^2}\}$ is bounded by $K$. Now take $n \in \mathbb{N}$ and let $J \subseteq I$ the set over which the $h_n$ is not bounded. Now for each $k \in \mathbb{N}$ we have $h_n^2 \geq k^2$ over $J$ and so $|J| \cdot k^2 < \int_J h_n^2 \leq \int_I h_n^2 \leq C^2$ i.e. $|J| \leq C^2/k^2$, $\forall k \in \mathbb{N}$, i.e. $|J| = 0$.

Let now $\mathcal{F} \subseteq L^2(I)$ a family of functions possibly not containing the trivial solution $f$ (better: not containing the strong closure of its span), and consider the closure of its spanned subspace $\tilde{\mathcal{F}} = \overline{(<\mathcal{F}>)}^{\|\cdot\|_{L^2}}$. Now $\tilde{\mathcal{F}}$ is hardly closed and $\Psi$ are (sequentially) continuous, so we want to prove that they have a minimum over $\tilde{\mathcal{F}}$.

**Proposition 2.4** *The functionals $\Psi$ have a minimum over $\tilde{\mathcal{F}}$.*

*Proof:* Define $J = I_{\tilde{\mathcal{F}}} + \Psi$ where $I_{\tilde{\mathcal{F}}}(u) = 0$ if $x \in \tilde{\mathcal{F}}$ and $I_{\tilde{\mathcal{F}}}(u) = +\infty$ if $x \notin \tilde{\mathcal{F}}$. Then $J$ is not trivial and $\inf J < +\infty$. For each $n \in \mathbb{N}$ there exists at least a $u_n \in L^2(I)$, and actually $u_n \in \tilde{\mathcal{F}}$, such that $|J(u_n) - \inf J| < \frac{1}{n}$. Observe that $J$ is itself coercive (with respect to the $\|\cdot\|_{L^2}$ metric) so that the sequence $\{u_n\}$ is uniformly bounded in $\tilde{\mathcal{F}}$. Eventually extracting a subsequence, it is possible to find $u \in L^2(I)$ such that $u_n \to u$ and $u$ has the property that $u \in \tilde{\mathcal{F}}$ by definition of $\tilde{\mathcal{F}}$ or because $J(u) < +\infty$ and $\Psi(u) = J(u) = \inf J = \inf_{\tilde{\mathcal{F}}} \Psi$.

**Note:** observe that in general $\Psi$ are not convex so $u$ may not be unique. Further note that if $f \in <\mathcal{F}>$ or $f \in \tilde{\mathcal{F}}$ the solution found in (2.4) is $f$. Note that these two last propositions prove that with an appropriate choice for $h(x)$ the error functionals have a minimum (not unique in general) and it is possible to find a minimization sequence. This is what we try to do numerically in next section.

## 3 The numerical solution

Following [4] one may try to find a gradient descending algorithm to minimize the cost functional. But after easy calculation one can see for instance that the two functional proposed above are not convex in general, and that in $L^2(I)$ it is easy to prove that $\nabla(\varphi_i) \equiv 0$. So we follow ([2, 3]) in order to find an algorithm to approximate the solution given in the previous section.

Consider now the algorithm in figure (1); This is a classical boosting algorithm, but

$$
\begin{aligned}
&(\rho_0, F_0) = ArgMin_{\{h \in \mathcal{F}, \rho \in \mathbb{R}\}} \Psi(\rho \cdot h, f) \\
&F_0 \leftarrow \rho_0 \cdot F_0 \\
&\textbf{for } m = 1...M \\
&\qquad \tilde{f} = f - F_{m-1} \\
&\qquad (\rho_m, h_m) = ArgMin_{\{h \in \mathcal{F}, \rho \in \mathbb{R}\}} \Psi(\rho \cdot h, \tilde{y}) \\
&\qquad F_m \leftarrow F_{m-1} + \rho_m \cdot h_m \\
&\textbf{endFor}
\end{aligned}
$$

Figure 1: **Gradient Boost algorithm**

the key point is the definition of the $ArgMin$ procedure that depends more on the definition of the functional.

First of all note that for any $\gamma \neq 0$, $\varphi(\gamma \cdot g) = \varphi(g)$, and it's easily found that minimizing the functional at this stage, we minimize just the distance component. For simplicity of notation view your family $\mathcal{F}$ as a matrix $h(a, i)$ where $h(a, \cdot) = s_a$. So recalling classical *greedy* algorithm these algorithms build in $M$ step the target function $\tilde{f}$ minimizing at each step the error:

$$\Lambda(\rho, a) = \sum_{i=1}^{N} \left( \tilde{y}_i - \rho \cdot \sum_j \delta_a^j s_i(t_j) \right)^2 = \sum_{i=1}^{N} (\tilde{y}_i - \rho \cdot h(a, i))^2 \qquad (5)$$

where $\tilde{y}$ is the residual at the chosen step. Now, even if $\Lambda(\rho, a)$ is not $\mathcal{C}^2$, it is $\mathcal{C}^2$ in $\rho$, an it is easily found that:

$$argmin_\rho(\Lambda(\rho, a)) = \frac{\sum_{i=1}^{N} h(a,i) \cdot \tilde{y}_i}{\sum_{i=1}^{N} h(a,i)^2} \tag{6}$$

An implementation for the global *ArgMin* procedure could be found in figure (2). Here

```
error  =  LBound
for g ∈ F_m
        ρ(g) = argmin_ρ(Λ(ρ))
        If φ(ỹ, ρ(g) · g) ≥ error
                error = φ(ỹ, ρ(g) · g)
                ρ_m = ρ(g)
                h_m = g
        endIf
        F_{m+1} ← F_m \ {h_m}
endFor
```

Figure 2: The *ArgMin* algorithm

$LBound = -1$ by default, but when increased it controls the minimum required correlation (e.g.: $LBound = 0$ means that solution at this step is good iff the correlation of the chose function with the residue is positive).